\documentclass{article}

\makeatletter
\def\input@path{{Styles/}}
\makeatother
\usepackage{iclr2027_conference,times}
\usepackage{amsmath,amssymb,booktabs,graphicx}
\usepackage{subcaption}
\usepackage{xcolor,siunitx}
\usepackage{colortbl}
\usepackage{float}
\usepackage{placeins}
\usepackage{needspace}
\usepackage{longtable}
\usepackage{algorithm,algpseudocode}
\usepackage{xurl}
\usepackage[hidelinks]{hyperref}
\definecolor{TableBest}{HTML}{205B8D}
\definecolor{TableSecond}{HTML}{287568}
\definecolor{TableMuted}{HTML}{555E68}
\definecolor{RankFirst}{HTML}{E1EDF7}
\definecolor{RankSecond}{HTML}{E2F0DF}
\definecolor{RankThird}{HTML}{FFF1D6}
\definecolor{TableGroupFill}{HTML}{E9EEF4}
\definecolor{TableGroupText}{HTML}{263D53}
\newcommand{\MainTableGroup}[2]{%
  \specialrule{0.4pt}{0pt}{0pt}%
  \multicolumn{#1}{@{}l}{\cellcolor{TableGroupFill}%
    \hspace{3pt}%
    \color{TableGroupText}\bfseries #2}\\%
  \specialrule{0.4pt}{0pt}{2pt}%
}

\newcommand{\MainTableValue}[2][0]{%
  \ifcase#1%
    \tablenum[table-format=1.3,mode=text]{#2}%
  \or%
    \cellcolor{RankFirst}{\bfseries
      \tablenum[table-format=1.3,mode=text,detect-weight=true]{#2}}%
  \or%
    \cellcolor{RankSecond}\underline{%
      \tablenum[table-format=1.3,mode=text]{#2}}%
  \or%
    \cellcolor{RankThird}%
      \tablenum[table-format=1.3,mode=text]{#2}%
  \fi%
}

\newcommand{\MainTableRankLegend}{%
  \colorbox{RankFirst}{\textbf{1st}},
  \colorbox{RankSecond}{\underline{2nd}}, and
  \colorbox{RankThird}{3rd} mark the three best values in each column
  across the reported model configurations; ranks use unrounded summaries.%
}

\newcommand{\TableValue}[2][0]{%
  \ifcase#1%
    \tablenum[table-format=1.4,mode=text]{#2}%
  \or%
    {\color{TableBest}\bfseries
      \tablenum[table-format=1.4,mode=text,detect-weight=true]{#2}}%
  \or%
    {\color{TableSecond}\underline{%
      \tablenum[table-format=1.4,mode=text]{#2}}}%
  \fi%
}
\newcommand{\TableMedian}[3][0]{%
  \begin{tabular}[t]{@{}c@{}}%
    \TableValue[#1]{#2}\\[-1.5pt]%
    {\scriptsize\color{TableMuted}%
      \if\relax\detokenize{#3}\relax
        \phantom{\ensuremath{\pm}\,0.0000}%
      \else
        \ensuremath{\pm}\,\tablenum[table-format=1.4,mode=text]{#3}%
      \fi}%
  \end{tabular}%
}

\renewcommand{\iclrruler}[1]{}

\title{RiverScan: Learning to Accumulate on River Networks}

\author{
He Zhu$^{1}$, Zhiyi Gao$^{1}$, Youhuan Yang$^{1}$, Han Quan$^{1}$, \& Ming Li$^{2}$\\
$^{1}$Central China Institute of Artificial Intelligence, Zhengzhou, China\\
$^{2}$Cheriton School of Computer Science, University of Waterloo, Ontario, Canada\\
\texttt{\{zhuhe,gaozhiyi,yangyouhuan,quanhan\}@zhongyuanai.cn}\\
\texttt{mli@uwaterloo.ca}
}

\iclrfinalcopy

\begin{document}

\maketitle
\fancyhead{}
\fancyhead[L]{Preprint}

\begin{abstract}
Accurate runoff prediction across large river basins requires models to capture heterogeneous local hydrological responses and their accumulation downstream. River-network connectivity provides a natural structure for this task, but using it effectively requires learning how to combine contributions from subcatchments of different sizes. We frame this challenge as learning to accumulate and introduce RiverScan, a spatial operator based on area-weighted selective state-space recurrence. RiverScan recursively combines local temporal representations with accumulated upstream states in topological order. Contributing-area fractions account for subcatchment size, while learned retention controls how upstream information persists downstream. Experiments on 608 stations in the Danube basin show that RiverScan outperforms the evaluated temporal and graph baselines in overall predictive performance. Structural controls and ablations demonstrate that composing heterogeneous subcatchment representations in a manner aligned with hydrological accumulation substantially improves runoff prediction, with the benefits of correct topology growing along the river-network hierarchy. These findings establish physically structured accumulation as an effective inductive bias for hydrological modeling and motivate finer-grained, fully distributed AI-based hydrological models.
\end{abstract}

\section{Introduction}
\label{sec:introduction}

\begin{figure}[b]
  \centering
  \includegraphics[width=\linewidth]{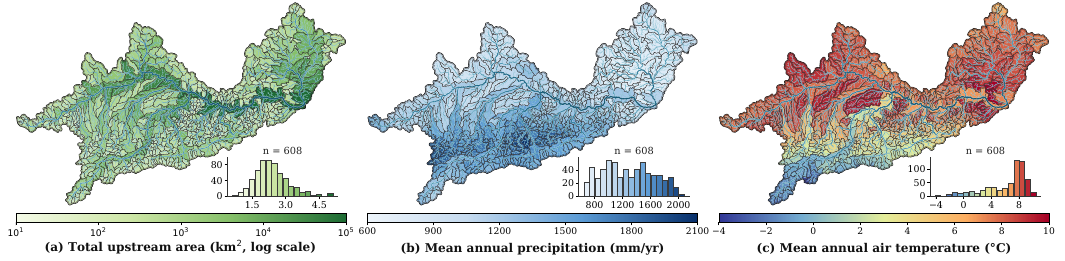}
  \caption{\textbf{Spatial extent and meteorological variation in the Danube basin.}
    (a) Total upstream contributing area $F_v$ (log scale).
    (b) Mean annual precipitation. (c) Mean annual air temperature.
    Insets summarize the distributions across 608 gauges.}
  \label{fig:hydrological-motivation}
\end{figure}

Accurate river discharge prediction supports flood warning and water resources management. It requires integrating responses from regions with different meteorological forcing, landscape properties, and contributing areas (Figure~\ref{fig:hydrological-motivation}). Shared LSTMs learn rainfall--runoff dynamics across diverse basins \citep{kratzert2018,kratzert2019,nearing2024global}, but catchment-averaged inputs compress the spatial variability within each basin. Encoding incremental catchments separately preserves local histories and requires learning how their representations combine into a basin-scale response.

River networks provide a natural structure for this composition. At each confluence, incoming tributaries carry the accumulated responses of their upstream catchments and combine with a local contribution. A learnable representation should therefore account for both the area it summarizes and how information persists downstream. We call this task \emph{learning to accumulate}.

In one-hop synchronous message passing, upstream coverage grows with layer depth \citep{kipf2017}. Degree normalization and learned attention \citep{velickovic2018} also do not explicitly account for the unequal contributing areas represented by tributary states. These considerations motivate an operator that combines completed tributary summaries using the spatial extent of each contributing region.

We introduce \emph{RiverScan}, an area-normalized selective recurrence on the river tree. Building on ordered graph recurrences \citep{moshe2020hydronets,thost2021} and tree state-space models \citep{xiao2024mambatree}, RiverScan combines local temporal encodings with completed upstream states in topological order. Contributing-area fractions determine their geometric shares, while selective transitions \citep{gu2024}, conditioned on tributary states and reach attributes, control information retention. The accumulated state admits an exact source--path decomposition into source-area fractions, local encodings, and retention products along river paths.

On 608 Danube gauges from LamaH-CE \citep{klingler2021}, RiverScan achieves the best overall performance among the evaluated temporal and graph baselines without observed discharge inputs. Graph comparisons hold local inputs and the LSTM architecture fixed. Structural controls and ablations demonstrate the predictive value of composing heterogeneous subcatchments according to river topology and contributing area. Gains from correct topology increase along the gauge hierarchy, showing that this organization becomes more valuable as more levels of upstream accumulation are involved.

\section{Related Work}
\label{sec:related-work}

\paragraph{Neural hydrology and river-network modeling.}
LSTM and EA-LSTM models learn rainfall--runoff dynamics across catchments from meteorological inputs and static attributes \citep{kratzert2018,kratzert2019}. \citet{nearing2024global} further show that LSTM-based forecasting outperforms the physics-based GloFAS system for extreme floods in ungauged watersheds. Spatial models differ in the quantities they propagate: HydroNets recursively combines local and upstream embeddings \citep{moshe2020hydronets}, \citet{vischer2025} route gridded runoff estimates, and differentiable Muskingum--Cunge models learn parameters within physical routing equations \citep{bindas2024improving}. These studies, however, do not provide a systematic evaluation across hundreds of connected gauges within a single large river basin. Evidence for the benefit of river connectivity also depends on the task. Using LamaH-CE, \citet{kirschstein2024} found no advantage from true topology in the tested models for six-hour-ahead forecasting from hourly discharge and meteorological histories. \citet{mosaffa2026} reported gains from graph routing modules for daily rainfall--runoff prediction in the Upper Danube. Together, these findings leave open when and how river-network topology provides an effective inductive bias for hydrological prediction. Our setting predicts daily discharge without observed discharge inputs and studies the accumulation of latent states defined over incremental catchments.

\paragraph{Ordered graph recurrence and tree state-space models.}
Tree-LSTM combines child states before updating a parent \citep{tai2015}, while DAGNN updates nodes according to a graph's partial order \citep{thost2021}. On visual tasks, MambaTree constructs input-dependent minimum spanning trees over image features for classification, detection, and segmentation, and expresses states as sums of source terms weighted by path products \citep{xiao2024mambatree}. RiverScan builds on ordered propagation by specifying an area-normalized combination rule for tributary states with known spatial extents. Its area ratios telescope to source-area fractions, while retention depends on the completed catchment state and reach attributes.

\paragraph{Selective state spaces for spatial hydrological modeling.}
Mamba conditions state-space parameters on sequence inputs to control information retention \citep{gu2024}. RiverMamba targets scalable, dense global discharge and flood forecasting, using space-filling-curve serialization and bidirectional Mamba blocks to capture large spatiotemporal contexts \citep{shamseddin2025rivermamba}. RiverScan addresses spatial composition in rainfall--runoff modeling: how heterogeneous information from incremental catchments of unequal area should accumulate and attenuate through successive confluences. This motivates an area-weighted recurrence on the supplied river tree, with each edge transition conditioned on the completed upstream state and reach attributes.

\section{Problem Setting}
\label{sec:problem-setting}

We study daily rainfall--runoff prediction on a gauge tree $G=(V,E)$ directed downstream. Let $\mathcal C(v)$ denote the immediate upstream gauges of $v$. The incremental catchment at $v$ is the region draining to $v$ after excluding their total catchments. Writing $a_v>0$ for its area and $F_v$ for the total contributing area at $v$ gives the recursive partition (Appendix~\ref{app:catchment-nesting})
\begin{equation}
  F_v=a_v+\sum_{u\in\mathcal C(v)} F_u.
  \label{eq:contributing-area}
\end{equation}

At date $t$, inputs comprise an $L$-day meteorological history $\mathbf x_{v,t-L+1:t}$ and static attributes $\mathbf s_v$ for each incremental catchment, together with reach attributes $\mathbf e_{uv}$ for each edge $(u,v)$. We predict same-day runoff depth $y_{v,t}$ at every gauge using meteorology through day $t$, without observed discharge inputs.

Following common rainfall--runoff practice \citep{lees2021}, we normalize observed daily discharge by the reported total drainage area $F_v^{\mathrm{obs}}$ to obtain $y_{v,t}$ in mm\,day$^{-1}$. The recursively accumulated area $F_v$ determines the area fractions used for spatial composition (Equation~\ref{eq:area-prior}). Unit conversions are given in Appendix~\ref{app:input-training}.

\section{RiverScan: Selective Accumulation on River Networks}
\label{sec:method}

\begin{figure}[t]
  \centering
  \includegraphics[width=\linewidth]{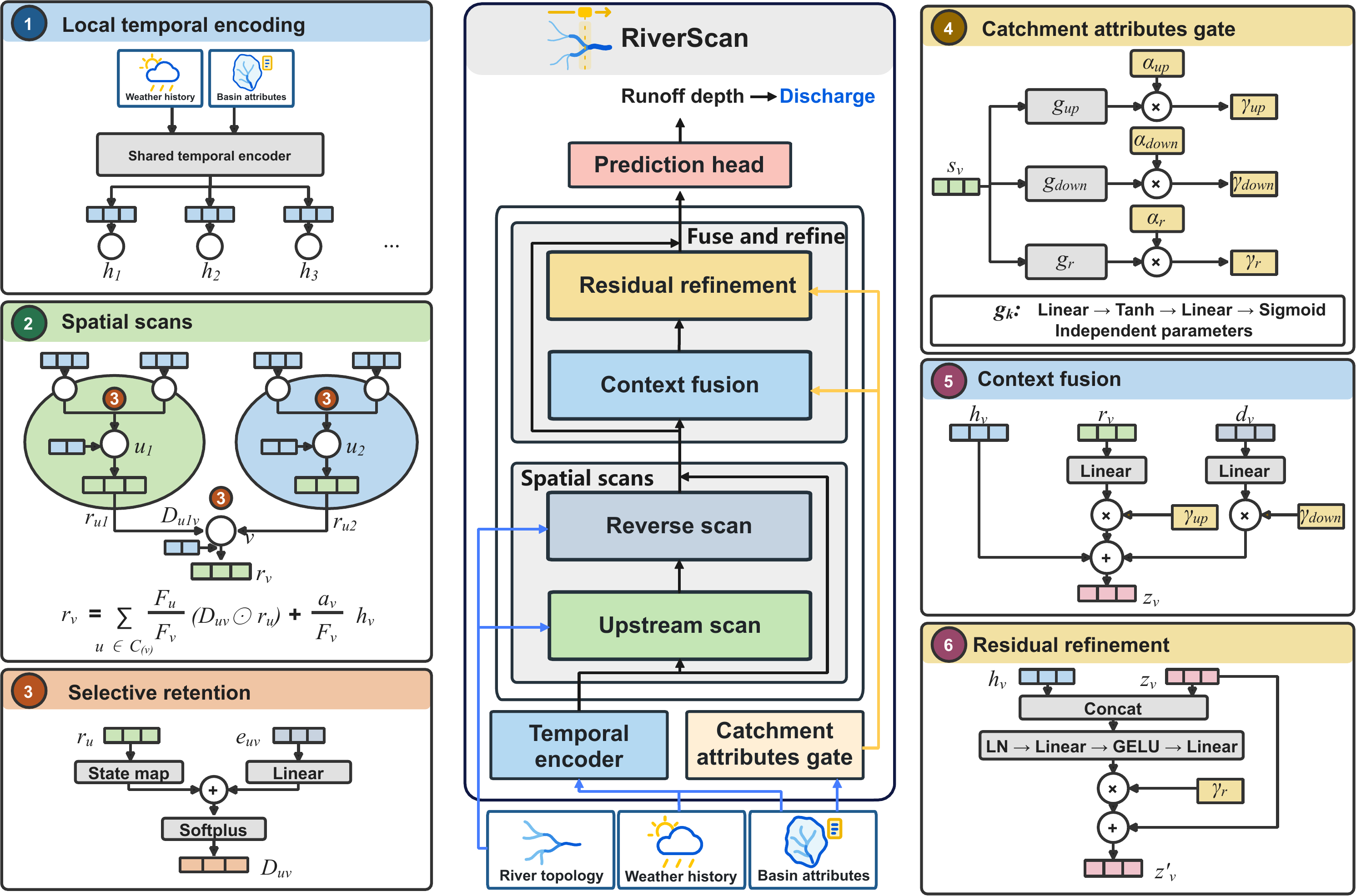}
  \caption{\textbf{RiverScan architecture.} Local temporal encodings (1) are composed through area-weighted upstream accumulation (2) with selective retention (3). An independent reverse scan supplies downstream context. Attribute-conditioned gates (4) regulate context fusion (5) and residual refinement (6) before runoff prediction. The central column summarizes the workflow.}
  \label{fig:architecture}
\end{figure}

RiverScan learns to accumulate heterogeneous local information into representations of complete contributing catchments. A shared temporal encoder preserves each incremental catchment's history, and an upstream recurrence combines these encodings according to river topology, contributing area, and learned retention. A separate downstream scan supplies context for the final prediction (Figure~\ref{fig:architecture}).

\subsection{Hydrological Basis for Area-Weighted Composition}
\label{sec:method-hydrological-prior}

Under linear routing, discharge at a target section combines routed inflows from its immediate upstream catchments with the local incremental contribution \citep{lohmann1996,mizukami2016}:
\begin{equation}
  Q_v(t)=\sum_{u\in\mathcal C(v)}\mathcal R_{uv}[Q_u](t)
      +Q_v^{\mathrm{loc}}(t).
  \label{eq:network-routing}
\end{equation}
Here $Q_v^{\mathrm{loc}}$ includes routing within the incremental catchment. The reach operator $\mathcal R_{uv}$ describes downstream transfer, for example through the impulse response $\mathcal R_{uv}[Q_u](t)=\int_0^t k_{uv}(\tau)Q_u(t-\tau)\,\mathrm d\tau$.

Expressing total and local discharge as runoff depths using $F_v$ and $a_v$ gives $q_v$ and $q_v^{\mathrm{loc}}$, respectively (unit conversions in Appendix~\ref{app:input-training}). Linearity then yields
\begin{equation}
  q_v(t)
  =\underbrace{\sum_{u\in\mathcal C(v)}\frac{F_u}{F_v}
       \mathcal R_{uv}[q_u](t)}_{\text{routed upstream contribution}}
   +\underbrace{\frac{a_v}{F_v}q_v^{\mathrm{loc}}(t)}_{\text{local contribution}}.
  \label{eq:area-prior}
\end{equation}
Thus the local and tributary fractions are $\ell_v=a_v/F_v$ and $w_{uv}=F_u/F_v$, with $\ell_v+\sum_{u\in\mathcal C(v)}w_{uv}=1$. These fractions follow from catchment geometry and depth normalization while allowing heterogeneous local runoff responses. RiverScan uses this composition rule to organize learned catchment representations.

\subsection{Selective Accumulation of Catchment States}
\label{sec:method-upstream}

\subsubsection{Local Temporal Encoding}
\label{sec:method-encoder}

A shared LSTM \citep{hochreiter1997} maps each incremental catchment's meteorological history and static attributes to a local state:
\begin{equation}
  \mathbf h_{v,t}
  =f_\theta\!\left(
    [\mathbf x_{v,\tau};\mathbf s_v]_{\tau=t-L+1}^{t}
  \right)\in\mathbb R^d.
  \label{eq:temporal}
\end{equation}
Each catchment retains its own inputs while sharing encoder parameters. The state $\mathbf h_{v,t}$ summarizes conditions over the local area $a_v$. All spatial operations below act at a fixed date, so we suppress $t$.

\subsubsection{Area-Weighted Upstream Recurrence}
\label{sec:method-recurrence}

Let $\mathbf r_u$ summarize the complete tributary catchment of area $F_u$. RiverScan combines the completed tributary states with the local encoding:
\begin{equation}
  \mathbf r_v
  =
  \sum_{u\in\mathcal C(v)}
  \underbrace{\frac{F_u}{F_v}}_{\text{area prior}}
  \underbrace{\overline{\mathbf A}_{uv}\mathbf r_u}_{\text{tributary transition}}
  +
  \underbrace{\frac{a_v}{F_v}\mathbf h_v}_{\text{local contribution}},
  \label{eq:upstream}
\end{equation}
where $\overline{\mathbf A}_{uv}=\operatorname{diag}(\mathbf D_{uv})$ retains information independently in each feature channel. The area fractions specify the spatial extent represented by each term, and the learned transition controls how much upstream information persists. The resulting $\mathbf r_v$ summarizes the contributing catchment at $v$ and becomes an input to the next downstream update.

This transition-plus-input structure follows state-space recurrence \citep{gu2024}, with multiple completed tributary states supplying the predecessors and $\ell_v\mathbf h_v$ supplying the local injection. The physical accumulation rule constrains how the learned features are composed.

\subsubsection{State- and Reach-Conditioned Retention}
\label{sec:method-retention}

Retention depends on both the completed tributary state and the connecting reach:
\begin{equation}
  \begin{aligned}
  \boldsymbol\delta_{uv}
    &=\operatorname{softplus}\!\left(
      W_s\operatorname{LN}(\mathbf r_u)
      +W_e\mathbf e_{uv}+\mathbf b
    \right),\\
  \mathbf D_{uv}
    &=\exp\!\left[
      -\boldsymbol\delta_{uv}\odot\exp(\boldsymbol\eta)
    \right].
  \end{aligned}
  \label{eq:retention}
\end{equation}
The reach attributes $\mathbf e_{uv}$ contain standardized distance, elevation difference, and slope. The parameters $W_s,W_e,\mathbf b,\boldsymbol\eta$ are shared across reaches, and layer normalization acts on the state entering the gate. Conditioning on $\mathbf r_u$ lets retention adapt to the accumulated upstream information. Each component of $\mathbf D_{uv}$ lies in $(0,1)$.

With $\mathbf A=-\operatorname{diag}(\exp\boldsymbol\eta)$, the transition has the diagonal exponential form used in selective state-space models \citep{gu2024}:
\begin{equation}
  \overline{\mathbf A}_{uv}
  =
  \exp\!\left[
    \operatorname{diag}(\boldsymbol\delta_{uv})\mathbf A
  \right].
  \label{eq:exponential-transition}
\end{equation}
Here $\boldsymbol\delta_{uv}$ controls feature retention across a reach at the current date. Temporal dependencies are represented by the local encoder.

\subsection{Accumulation Order and Source--Path Decomposition}
\label{sec:method-decomposition}

The upstream scan starts at headwaters, where $F_v=a_v$ and $\mathbf r_v=\mathbf h_v$. It proceeds in topological order, completing each tributary state before computing its outgoing transition. One traversal incorporates the full upstream subtree, and summation makes the update invariant to tributary ordering. Independent branches, dates, and batch elements can be processed in parallel. State-dependent transitions require sequential evaluation along each river path (Appendix~\ref{app:scan-computation}).

Let $S(v)$ contain $v$ and its upstream nodes, and let $P(i,v)$ denote the unique directed path from $i$ to $v$. Define the path retention $\mathbf K_{i\to v}=\bigodot_{e\in P(i,v)}\mathbf D_e$, with $\mathbf K_{v\to v}=\mathbf1$. Expanding the recurrence gives
\begin{equation}
  \mathbf r_v
    =\sum_{i\in S(v)}\frac{a_i}{F_v}
      \bigl(\mathbf K_{i\to v}\odot\mathbf h_i\bigr).
  \label{eq:source-decomposition-main}
\end{equation}
This identity holds for the state-dependent retentions realized in each forward pass (proof in Appendix~\ref{app:source-proof}). Each source contributes its local encoding, weighted by its fraction of the target area and the retention accumulated along its river path. With unit retention on every edge, the state reduces to the area-weighted mean $F_v^{-1}\sum_{i\in S(v)}a_i\mathbf h_i$. This decomposition connects the accumulated representation to the heterogeneous regions it summarizes.

\subsection{Context Fusion and Runoff Prediction}
\label{sec:method-readout}

\paragraph{Downstream context.}
An independent reverse scan incorporates encoded conditions along the downstream chain. Let $p(v)$ denote the next downstream gauge. Starting at the outlet, the scan computes
\begin{equation}
  \mathbf d_v=
  \begin{cases}
    \mathbf h_v, & p(v)\text{ does not exist},\\
    \mathbf h_v+
    \mathbf D^\downarrow_{p(v)\to v}\odot\mathbf d_{p(v)},
      & \text{otherwise}.
  \end{cases}
  \label{eq:downstream}
\end{equation}
Its retention uses the form in Equation~\ref{eq:retention}, with independent parameters conditioned on $\mathbf d_{p(v)}$ and the attributes of $(v,p(v))$. Both local and downstream terms enter with unit area coefficients. The state $\mathbf d_v$ supplies statistical context from the encoded inputs.

\begin{samepage}
\paragraph{Gated fusion and refinement.}
Static catchment attributes control the contribution of each spatial branch and the refinement module. For $k\in\{\uparrow,\downarrow,r\}$, let $\gamma_{k,v}=\alpha_k g_k(\mathbf s_v)$, where each $g_k$ is a separately parameterized Linear--Tanh--Linear--Sigmoid network and $\alpha_k$ is a learned global scale. Prediction proceeds through
\begin{align}
    \mathbf z_v
      &=\mathbf h_v
      +\gamma_{\uparrow,v}P_\uparrow(\mathbf r_v)
      +\gamma_{\downarrow,v}P_\downarrow(\mathbf d_v),
      \label{eq:fusion}\\
    \mathbf z'_v
      &=\mathbf z_v+
      \gamma_{r,v}F_\psi([\mathbf h_v;\mathbf z_v]),
      \label{eq:refine}\\
    \widehat y_v
      &=\operatorname{Head}\!\left(
      \operatorname{Dropout}(\mathbf z'_v)\right).
      \label{eq:prediction}
\end{align}
\end{samepage}
The affine maps $P_\uparrow$ and $P_\downarrow$ adapt the spatial states for fusion. The nodewise refinement $F_\psi$ is a LayerNorm--Linear--GELU--Linear network applied after both scans. Reach retention governs representation formation, while static gates govern its use in prediction. All three scales $\alpha_k$ start at zero, initializing the predictor on its local temporal path.

\section{Experimental Design}
\label{sec:experiments}

\paragraph{Data and task.}
\label{sec:experimental-data}
We use the Danube (LAMA-B) network from LamaH-CE \citep{klingler2021}, with 608 gauges and 607 directed edges. Our LSTM experiments follow the daily rainfall--runoff modeling framework of \citet{kratzert2018,kratzert2019}, learning from meteorological histories with parameters shared across catchments. Training, validation, and test periods are 1981--2000, 2000--2003, and 2003--2017; exact dates, inputs, and training settings are in Appendix~\ref{app:experimental-details}. Each run uses its lowest-validation-loss checkpoint, with test metrics computed after returning discharge to mm\,day$^{-1}$.

\paragraph{Comparisons.}
\label{sec:experimental-models}
Graph comparisons hold the LSTM architecture and local incremental-catchment inputs fixed to study spatial aggregation and accumulation. Table~\ref{tab:main} compares RiverScan with eight temporal and graph baselines. Table~\ref{tab:components} reports connectivity controls and matched component removals. The LSTM reference uses full-catchment forcing.
\label{sec:experimental-controls}

\paragraph{Evaluation.}
\label{sec:experimental-evaluation}
We average results across three training seeds for each configuration. NSE and KGE are unbounded below, so extreme negative scores at a few gauges can dominate arithmetic averages. We therefore use station medians, following common practice in large-sample hydrology \citep{kratzert2019}, and report CDF areas on $[0,1]$ (lower is better). Results cover all 608 gauges, 314 headwaters, and 294 non-headwaters. Seed aggregation, paired differences, win rates, and bootstrap definitions are specified in Appendix~\ref{app:evaluation-details}.

\section{Results and Analysis}
\label{sec:results}

\subsection{Predictive Value of Spatial Composition}
\label{sec:results-topology}

\begin{table}[!htbp]
\centering
\caption{Danube performance by network position, with groups fixed on the observed tree. Higher medians and lower CDF areas are better. \protect\MainTableRankLegend}
\label{tab:main}
\fontsize{8}{9.5}\selectfont
\setlength{\tabcolsep}{1.2pt}
\renewcommand{\arraystretch}{1.28}
\begin{tabular*}{\linewidth}{@{\extracolsep{\fill}}l*{4}{c}@{\hspace{2pt}}*{4}{c}@{\hspace{2pt}}*{4}{c}@{}}
\toprule
& \multicolumn{4}{c}{\textbf{All stations} ($n=608$)} & \multicolumn{4}{c}{\textbf{Headwaters} ($n=314$)} & \multicolumn{4}{c}{\textbf{Non-headwaters} ($n=294$)} \\
\cmidrule(lr){2-5}\cmidrule(lr){6-9}\cmidrule(lr){10-13}
Model & \multicolumn{2}{c}{NSE} & \multicolumn{2}{c}{KGE} & \multicolumn{2}{c}{NSE} & \multicolumn{2}{c}{KGE} & \multicolumn{2}{c}{NSE} & \multicolumn{2}{c}{KGE} \\
\cmidrule(lr){2-3}\cmidrule(lr){4-5}\cmidrule(lr){6-7}\cmidrule(lr){8-9}\cmidrule(lr){10-11}\cmidrule(lr){12-13}
& Med $\uparrow$ & CDF $\downarrow$ & Med $\uparrow$ & CDF $\downarrow$ & Med $\uparrow$ & CDF $\downarrow$ & Med $\uparrow$ & CDF $\downarrow$ & Med $\uparrow$ & CDF $\downarrow$ & Med $\uparrow$ & CDF $\downarrow$ \\
\MainTableGroup{13}{Per-station temporal baselines}
LSTM & \MainTableValue{0.657} & \MainTableValue{0.391} & \MainTableValue{0.744} & \MainTableValue{0.297} & \MainTableValue{0.602} & \MainTableValue{0.432} & \MainTableValue{0.701} & \MainTableValue{0.335} & \MainTableValue{0.707} & \MainTableValue{0.347} & \MainTableValue{0.783} & \MainTableValue{0.256} \\
EA-LSTM & \MainTableValue{0.636} & \MainTableValue{0.406} & \MainTableValue{0.718} & \MainTableValue{0.318} & \MainTableValue{0.619} & \MainTableValue{0.422} & \MainTableValue{0.698} & \MainTableValue{0.335} & \MainTableValue{0.651} & \MainTableValue{0.388} & \MainTableValue{0.736} & \MainTableValue{0.299} \\
Mamba & \MainTableValue{0.578} & \MainTableValue{0.473} & \MainTableValue{0.661} & \MainTableValue{0.379} & \MainTableValue{0.538} & \MainTableValue{0.518} & \MainTableValue{0.621} & \MainTableValue{0.420} & \MainTableValue{0.630} & \MainTableValue{0.426} & \MainTableValue{0.698} & \MainTableValue{0.335} \\
\addlinespace[3pt]
\MainTableGroup{13}{Graph neural network baselines}
GraphSAGE (2L) & \MainTableValue[3]{0.690} & \MainTableValue[2]{0.346} & \MainTableValue[3]{0.763} & \MainTableValue[2]{0.272} & \MainTableValue[3]{0.662} & \MainTableValue[2]{0.374} & \MainTableValue[1]{0.737} & \MainTableValue[2]{0.303} & \MainTableValue[3]{0.714} & \MainTableValue[3]{0.316} & \MainTableValue{0.790} & \MainTableValue[2]{0.239} \\
GCN (2L) & \MainTableValue{0.684} & \MainTableValue{0.361} & \MainTableValue{0.755} & \MainTableValue{0.284} & \MainTableValue{0.648} & \MainTableValue{0.389} & \MainTableValue{0.723} & \MainTableValue{0.322} & \MainTableValue{0.709} & \MainTableValue{0.330} & \MainTableValue[2]{0.794} & \MainTableValue[3]{0.244} \\
GCNII (2L) & \MainTableValue{0.652} & \MainTableValue{0.387} & \MainTableValue{0.679} & \MainTableValue{0.348} & \MainTableValue{0.620} & \MainTableValue{0.413} & \MainTableValue{0.644} & \MainTableValue{0.380} & \MainTableValue{0.685} & \MainTableValue{0.359} & \MainTableValue{0.711} & \MainTableValue{0.313} \\
GAT (4L) & \MainTableValue{0.635} & \MainTableValue{0.435} & \MainTableValue{0.750} & \MainTableValue{0.303} & \MainTableValue{0.624} & \MainTableValue{0.447} & \MainTableValue[3]{0.732} & \MainTableValue{0.321} & \MainTableValue{0.652} & \MainTableValue{0.422} & \MainTableValue{0.764} & \MainTableValue{0.284} \\
DAGNN (2L) & \MainTableValue[2]{0.699} & \MainTableValue[3]{0.347} & \MainTableValue[2]{0.766} & \MainTableValue[3]{0.276} & \MainTableValue[2]{0.666} & \MainTableValue[3]{0.382} & \MainTableValue{0.729} & \MainTableValue[3]{0.304} & \MainTableValue[2]{0.736} & \MainTableValue[2]{0.310} & \MainTableValue[3]{0.794} & \MainTableValue{0.245} \\
\midrule
\textbf{RiverScan} & \MainTableValue[1]{0.709} & \MainTableValue[1]{0.332} & \MainTableValue[1]{0.773} & \MainTableValue[1]{0.266} & \MainTableValue[1]{0.675} & \MainTableValue[1]{0.371} & \MainTableValue[2]{0.736} & \MainTableValue[1]{0.299} & \MainTableValue[1]{0.743} & \MainTableValue[1]{0.290} & \MainTableValue[1]{0.806} & \MainTableValue[1]{0.230} \\
\bottomrule
\end{tabular*}
\end{table}

In every metric and station group in Table~\ref{tab:main}, the top three models all incorporate spatial information through different composition operators. This pattern highlights the predictive relevance of spatial heterogeneity at the basin scales represented in LamaH-CE and the benefits of explicitly modeling spatial relationships.

RiverScan achieves the highest median NSE/KGE and lowest CDF areas both overall and at non-headwater gauges (Table~\ref{tab:main}). Its overall medians are $0.709$/$0.773$, compared with $0.699$/$0.766$ for the ordered DAGNN baseline and $0.657$/$0.744$ for the full-catchment LSTM. Graph comparisons share local inputs and the LSTM architecture, supporting the predictive value of the spatial composition operator.

\begin{figure}[!htbp]
  \centering
  \includegraphics[width=\linewidth]{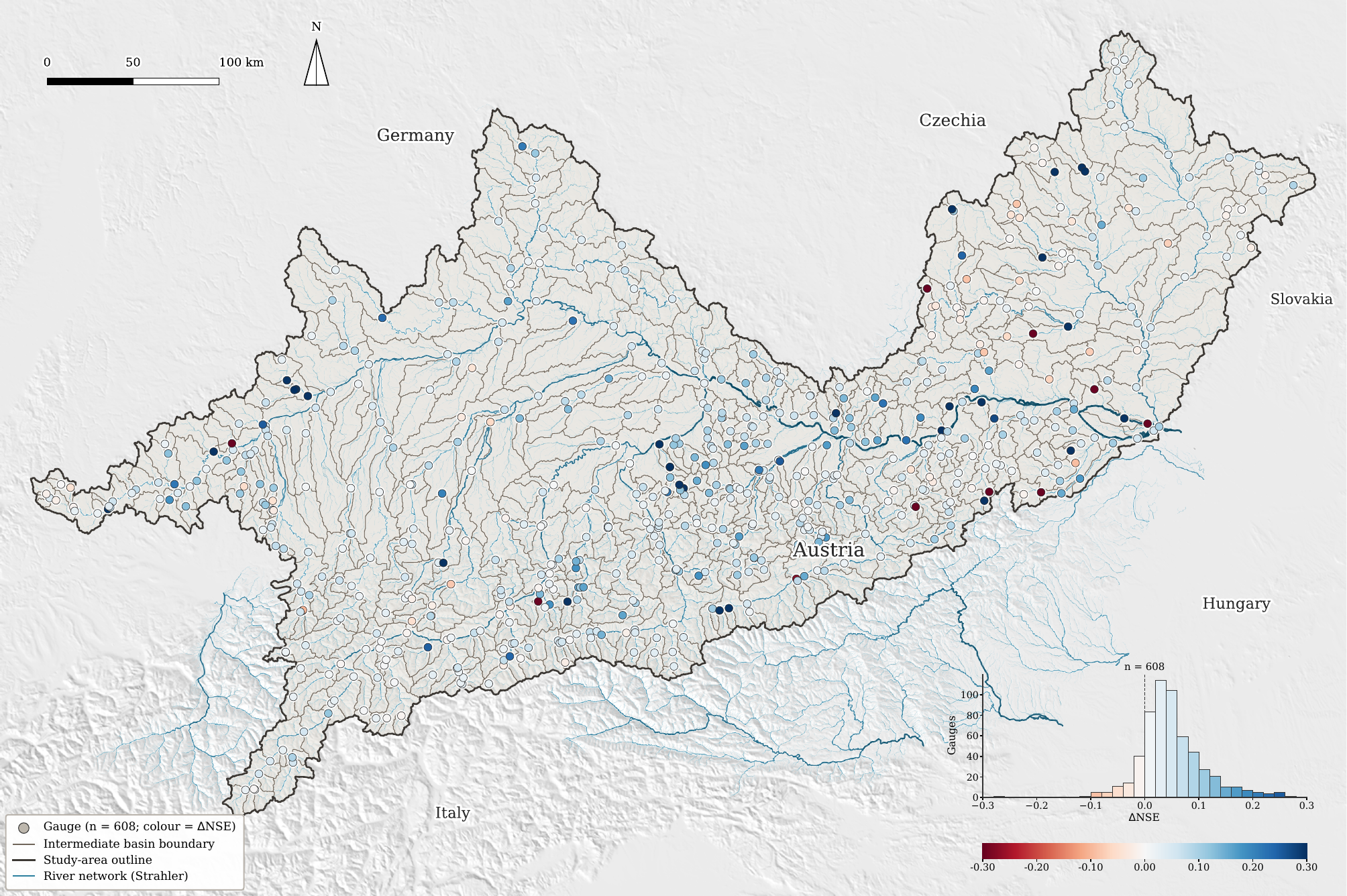}
  \caption{\textbf{Spatial gains over LSTM.} Gauge-level NSE differences use three-seed averages (Equation~\ref{eq:paired-gauge-nse}). Blue/red denote gains/losses; the inset shows all 608 differences with a dashed zero reference.}
  \label{fig:spatial-gain}
\end{figure}

Figure~\ref{fig:spatial-gain} maps gains over the full-catchment LSTM, a comparison that also changes input resolution.

\FloatBarrier
\subsection{Topology Gains Grow with Upstream Accumulation}
\label{sec:results-scale}

The topology controls retain station inputs and targets while altering network organization (Table~\ref{tab:components}). Permutation reassigns stations to tree positions and recomputes contributing areas. It yields relative paired NSE changes of $-2.00\%$ overall and $-3.74\%$ at non-headwaters, compared with $-5.46\%$ and $-7.07\%$ when edges are removed. Across seeds, RiverScan's mean win rates over permutation are $62.7\%$ overall and $76.8\%$ at non-headwaters.

\begin{figure}[!htbp]
  \centering
  \includegraphics[width=\linewidth]{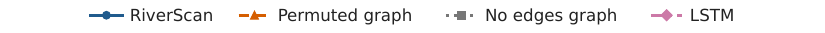}
  \par\vspace{4pt}
  \begin{subfigure}[t]{0.315\linewidth}
    \centering
    \includegraphics[width=\linewidth]{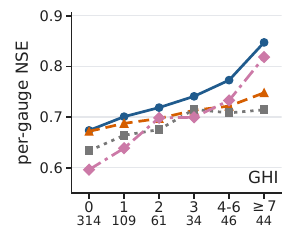}
    \caption{Median test NSE.}
    \label{fig:alignment-a}
  \end{subfigure}\hspace{0.027\linewidth}%
  \begin{subfigure}[t]{0.315\linewidth}
    \centering
    \includegraphics[width=\linewidth]{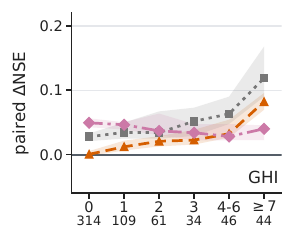}
    \caption{Median paired gain, 95\% CI.}
    \label{fig:alignment-b}
  \end{subfigure}\hspace{0.027\linewidth}%
  \begin{subfigure}[t]{0.315\linewidth}
    \centering
    \includegraphics[width=\linewidth]{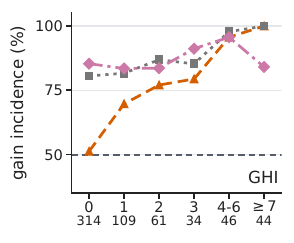}
    \caption{Positive-gain rate.}
    \label{fig:alignment-c}
  \end{subfigure}
  \caption{\textbf{Alignment benefit by gauge hierarchy level (GHI).}
    Gauge counts appear below the bins defined on the observed tree.
    RiverScan uses the observed network, and LSTM uses full-catchment inputs.
    Each gauge's NSE is averaged over three seeds before computing the panel statistics.
    Colours in (b,c) identify the reference in $\Delta\mathrm{NSE}=\mathrm{NSE}_{\mathrm{RiverScan}}-\mathrm{NSE}_{\mathrm{ref}}$.
    Bootstrap resampling treats gauges as independent.}
  \label{fig:alignment}
\end{figure}

Figure~\ref{fig:alignment} locates this benefit along the gauge hierarchy. GHI counts edges on the longest path from a headwater gauge to the target gauge, using the observed tree. With NSE first averaged across seeds and expressed as a percentage, the median paired gain over permutation grows from $0.05$ percentage points at leaves to $8.24$ at GHI $\ge7$, where all 44 gauges improve.

The full-catchment LSTM also improves with GHI (Figure~\ref{fig:alignment}a), consistent with a classical hydrological scaling argument. As \citet[p.~1]{peel2020} note, ``Whereas, at larger spatial and temporal scales, process complexity generally reduces through averaging smaller scale complexities.'' Spatial aggregation can therefore produce simpler outlet responses despite heterogeneous local processes, helping explain the strong performance of catchment-level LSTMs in larger basins. RiverScan's growing advantage over the permuted control shows that preserving and explicitly composing heterogeneous local histories provides additional predictive value as more upstream levels contribute.

\FloatBarrier
\subsection{Contributions of Area Weighting and Context}
\label{sec:results-components}

All five matched component removals yield negative relative paired NSE changes in both groups (Table~\ref{tab:components}). Removing area weighting produces the largest declines, $-1.62\%$ overall and $-2.23\%$ at non-headwaters. Median KGE also falls from $0.773$ to $0.764$ overall and from $0.806$ to $0.788$ at non-headwaters. This supports combining tributary states according to the contributing areas they summarize.

\begingroup
\begin{table}[!htb]
\centering
\caption{Structural controls and matched component ablations. $\Delta$NSE is the paired median change relative to the full-model median, computed within each seed. Win is the percentage of gauges where RiverScan has higher NSE. Seed summaries are averaged over three seeds; best medians at the reported precision are bold, including ties.
For $\Delta$NSE and Win, \colorbox{RankFirst}{\textbf{1st}}, \colorbox{RankSecond}{\underline{2nd}}, and \colorbox{RankThird}{3rd} mark the top three per column, ranked by $|\Delta\mathrm{NSE}|$ and Win, respectively.}
\label{tab:components}
\fontsize{8}{9.4}\selectfont
\setlength{\tabcolsep}{1pt}
\renewcommand{\arraystretch}{1.15}
\renewcommand{\MainTableGroup}[2]{%
  \specialrule{0.4pt}{0pt}{0pt}%
  \rowcolor{TableGroupFill}[0pt][0pt]%
  \multicolumn{#1}{@{}l@{}}{\color{TableGroupText}\bfseries #2}\\%
  \specialrule{0.4pt}{0pt}{0pt}%
}
\newcommand{\AblationRankFirst}[1]{\cellcolor{RankFirst}\textbf{#1}}
\newcommand{\AblationRankSecond}[1]{\cellcolor{RankSecond}\underline{#1}}
\newcommand{\AblationRankThird}[1]{\cellcolor{RankThird}#1}
\begin{tabular*}{\linewidth}{@{\extracolsep{\fill}}>{\raggedright\arraybackslash}p{1.50in}*{4}{c}@{\hspace{4pt}}*{4}{c}@{}}
\toprule
& \multicolumn{4}{c}{\textbf{All stations} ($n=608$)} & \multicolumn{4}{c}{\textbf{Non-headwaters} ($n=294$)} \\
\cmidrule(lr){2-5}\cmidrule(lr){6-9}
Variant & NSE med & KGE med & $\Delta$NSE (\%) & Win (\%) & NSE med & KGE med & $\Delta$NSE (\%) & Win (\%) \\
\MainTableGroup{9}{Full model}
\textbf{RiverScan} & \textbf{0.709} & \textbf{0.773} & --- & --- & \textbf{0.743} & 0.806 & --- & --- \\
\addlinespace[2pt]
\MainTableGroup{9}{A. Topology}
\hspace*{0.8em}-- Permuted & 0.693 & 0.766 & \AblationRankThird{-2.00} & 62.7 & 0.714 & 0.782 & \AblationRankThird{-3.74} & \AblationRankThird{76.8} \\
\hspace*{0.8em}-- No edges & 0.660 & 0.748 & \AblationRankSecond{-5.46} & \AblationRankSecond{77.4} & 0.682 & 0.766 & \AblationRankFirst{-7.07} & \AblationRankFirst{84.4} \\
\addlinespace[2pt]
\MainTableGroup{9}{B. Matched component ablations}
\hspace*{0.8em}w/o refinement & 0.691 & 0.770 & -1.59 & \AblationRankThird{64.3} & 0.735 & \textbf{0.808} & -0.88 & 60.9 \\
\hspace*{0.8em}w/o downstream scan & 0.694 & 0.767 & -1.21 & 61.6 & 0.737 & 0.805 & -0.40 & 55.4 \\
\hspace*{0.8em}w/o area weighting & 0.694 & 0.764 & -1.62 & 62.2 & 0.718 & 0.788 & -2.23 & 70.5 \\
\hspace*{0.8em}w/o gate-input normalization & 0.704 & 0.766 & -0.20 & 52.3 & 0.740 & 0.800 & -0.07 & 51.2 \\
\hspace*{0.8em}w/o retention edge features & 0.705 & 0.770 & -0.52 & 56.0 & \textbf{0.743} & 0.802 & -0.32 & 52.9 \\
\addlinespace[2pt]
\MainTableGroup{9}{C. Encoder baseline}
LSTM (full-catchment inputs) & 0.657 & 0.744 & \AblationRankFirst{-5.97} & \AblationRankFirst{81.2} & 0.707 & 0.783 & \AblationRankSecond{-4.91} & \AblationRankSecond{80.6} \\
\bottomrule
\end{tabular*}
\end{table}
\endgroup

Refinement and downstream context provide additional gains. Removing refinement gives relative paired NSE changes of $-1.59\%$ overall and $-0.88\%$ at non-headwaters, while removing the downstream scan gives $-1.21\%$ and $-0.40\%$. Removing gate-input normalization or retention edge features produces smaller paired NSE declines.

\FloatBarrier
\Needspace{7\baselineskip}
\subsection{Accumulated States Track Contributing Regions}
\label{sec:results-state-composition}

We probe the completed upstream state $\mathbf r_v$ before projection, static gating, fusion, and refinement (Figure~\ref{fig:architecture}). Its upstream and local components satisfy $\mathbf r_v=\mathbf r_v^{\mathrm{up}}+\mathbf r_v^{\mathrm{local}}$, with $\mathbf r_v^{\mathrm{local}}=(a_v/F_v)\mathbf h_v$. Signed projection shares $p_{\mathrm{up},v}$ and $p_{\mathrm{local},v}=1-p_{\mathrm{up},v}$ quantify their contributions along the direction of $\mathbf r_v$ (Appendix~\ref{app:state-composition}).

\begingroup
\begin{figure}[!htbp]
  \centering
  \begin{minipage}[b]{0.49\linewidth}
    \centering
    \includegraphics[width=\linewidth]{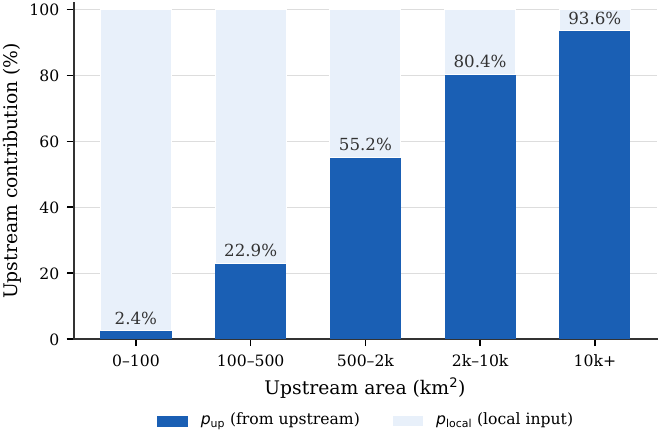}
    \par\smallskip
    {\small\textbf{(a) State composition by upstream area}}
  \end{minipage}\hfill
  \begin{minipage}[b]{0.49\linewidth}
    \centering
    \includegraphics[width=\linewidth]{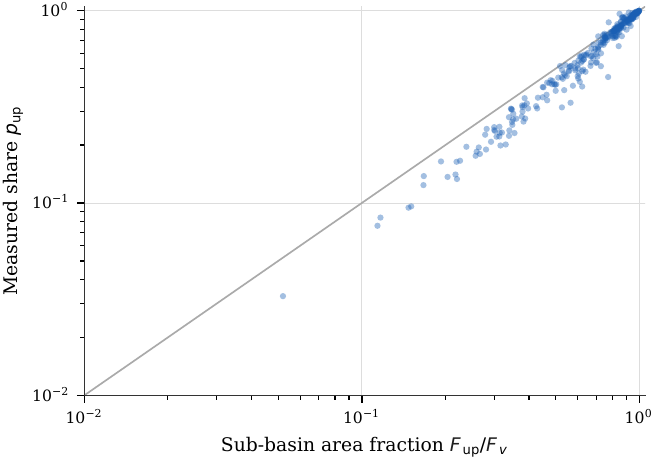}
    \par\smallskip
    {\small\textbf{(b) Area fraction vs. upstream state share}}
  \end{minipage}
  \caption{\textbf{Spatial composition of accumulated catchment representations.} Dimensionless projection shares (Equation~\ref{eq:state-projection-shares}) are computed after accumulation and before fusion, then averaged over three seeds. (a) Mean signed upstream (dark blue) and local (light blue) shares by $F_{\mathrm{up}}$ (km$^2$). (b) Gauge-level $p_{\mathrm{up}}$ versus geometric $F_{\mathrm{up}}/F_v$ on log--log axes, with a grey identity line.}
  \label{fig:state-composition}
\end{figure}
\endgroup

Mean $p_{\mathrm{up}}$ increases from $2.4\%$ to $93.6\%$ across upstream-area groups (Figure~\ref{fig:state-composition}a), showing a shift toward tributary-dominated representations. The positive gauge-level shares in Figure~\ref{fig:state-composition}b increase with $F_{\mathrm{up}}/F_v$, where $F_{\mathrm{up}}=\sum_{u\in\mathcal C(v)} F_u$, and generally lie below the identity line. These shares reflect area weights, learned retention, and the magnitudes and directions of regional encodings. The pattern is consistent with the source--path decomposition (Equation~\ref{eq:source-decomposition-main}), in which area determines each source's geometric weight and retention modulates its path contribution.

\FloatBarrier
\section{Limitations}
\label{sec:limitations}

Our evaluation is limited to LamaH-CE, whose paired incremental- and full-catchment data support the spatial comparisons studied here. As more paired incremental- and full-catchment data become available, we will assess the generality of our findings across a broader range of river basins. We focus on how subcatchment composition captures basin-scale spatial heterogeneity, using historical meteorological forcings to avoid additional uncertainty from weather-forecast products. Multi-day-ahead runoff forecasting therefore remains outside the scope of this study.

\section{Discussion and Conclusion}
\label{sec:discussion}

We introduced RiverScan, which learns to accumulate heterogeneous subcatchment representations through area-weighted selective recurrence on river networks. Experiments across 608 Danube gauges demonstrate that composing these representations in a manner aligned with hydrological accumulation substantially improves runoff prediction, with the benefits of correct topology growing along the river-network hierarchy. These findings establish physically structured accumulation as an effective inductive bias for learning basin-scale responses from heterogeneous local dynamics. Future work will explore finer-grained, fully distributed AI-based hydrological models, extending this principle toward more detailed spatial representations of hydrological processes.

\clearpage
\bibliographystyle{Styles/iclr2027_conference}

\begingroup
\interlinepenalty=10000
\bibliography{references}
\endgroup

\clearpage
\appendix
\section{Analysis and Implementation of RiverScan}
\label{app:method-analysis}

\subsection{Nested Catchments and Incremental Regions}
\label{app:catchment-nesting}

\begin{figure}[H]
  \centering
  \includegraphics[width=\linewidth]{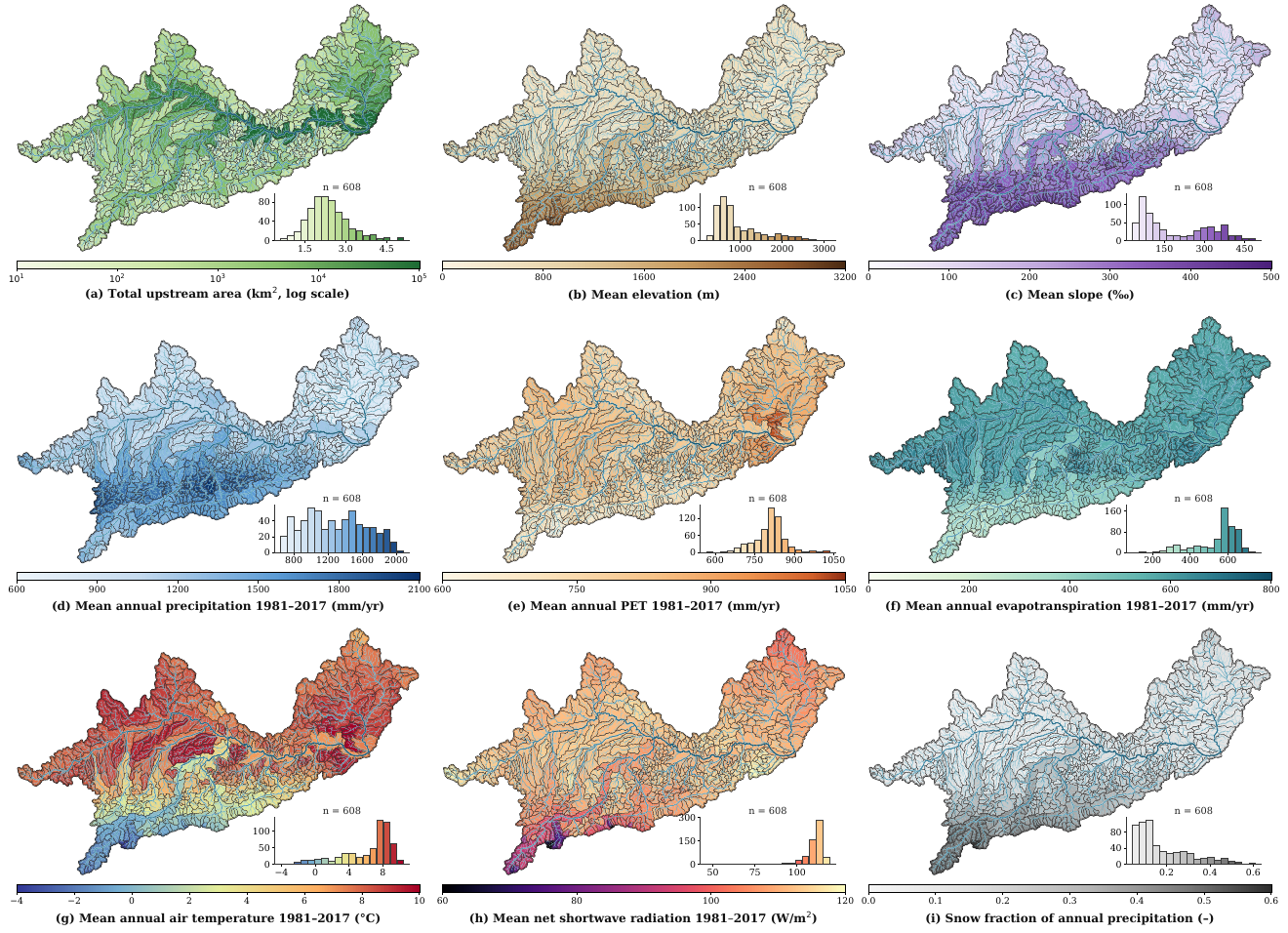}
  \caption{\textbf{Topographic and hydroclimatic characteristics of the Danube study domain.}
    (a--c) Total upstream contributing area (log scale), mean elevation, and mean slope.
    (d--h) 1981--2017 climatologies of precipitation, potential evapotranspiration (PET), evapotranspiration, air temperature, and net shortwave radiation.
    (i) Snow fraction of annual precipitation.
    Insets summarize the distributions across 608 gauges.}
  \label{fig:domain-overview}
\end{figure}

Total contributing catchments $\mathcal B_v$ are nested along river networks: if gauge $u$ is upstream of $v$, then $\mathcal B_u\subseteq\mathcal B_v$. Figure~\ref{fig:domain-overview}(a) shows their upstream areas across the Danube network. To count each source region once, define the incremental region by
\begin{equation}
  \mathcal I_v=\mathcal B_v\setminus
    \bigcup_{u\in\mathcal C(v)}\mathcal B_u,
  \qquad
  \mathcal B_v=\mathcal I_v\,\dot\cup\,
    \mathop{\dot\bigcup}_{u\in\mathcal C(v)}\mathcal B_u.
  \label{eq:catchment-partition}
\end{equation}
The unions have disjoint interiors under the river-tree assumption. With $a_v=\operatorname{area}(\mathcal I_v)$ and $F_v=\operatorname{area}(\mathcal B_v)$, this gives $F_v=a_v+\sum_{u\in\mathcal C(v)}F_u$. RiverScan encodes local meteorology over $\mathcal I_v$ and combines it with completed tributary summaries of $\mathcal B_u$, using area fractions $a_v/F_v$ and $F_u/F_v$.

Figure~\ref{fig:domain-overview}(b--i) also reveals meteorological heterogeneity across the contributing regions. The higher-elevation southern subcatchments are generally cooler and wetter, with larger snow fractions, while warmer, lower-elevation regions show greater potential evapotranspiration (b,d,e,g,i). Actual evapotranspiration and net shortwave radiation exhibit further spatial contrasts (f,h). Thus, the river network integrates regions with different water and energy inputs. Catchment-wide averages compress these contrasts, motivating local temporal encoding of incremental regions before their representations are accumulated downstream.

\Needspace{7\baselineskip}
\subsection{Source--Path Decomposition and Corollaries}
\label{app:source-proof}

Consider a finite directed tree, or a forest treated componentwise, with edges oriented downstream and at most one downstream successor per node. Let $S(v)$ contain $v$ and its upstream ancestors, and let $P(i,v)$ denote the unique directed path from $i$ to $v$. The local areas $a_v>0$ correspond to disjoint incremental catchments. The recurrence uses the recursively summed area $F_v=\sum_{i\in S(v)}a_i$. All statements below concern the upstream scan at one date.

Define the realized retention along a source-to-target path by
\begin{equation}
  \mathbf K_{i\to v}
    =\bigodot_{e\in P(i,v)}\mathbf D_e,\qquad
  \mathbf K_{v\to v}=\mathbf 1.
  \label{eq:path-retention}
\end{equation}
\textbf{Proposition 1 (area-normalized source decomposition).}
For any forward pass, the upstream recurrence in Equation~\ref{eq:upstream} satisfies
\begin{equation}
  \mathbf r_v
    =\sum_{i\in S(v)}\frac{a_i}{F_v}
      \bigl(\mathbf K_{i\to v}\odot\mathbf h_i\bigr).
  \label{eq:source-decomposition}
\end{equation}

\textit{Proof.} For a fixed forward pass, retain the computed values $\{\mathbf D_e\}$. At a headwater $v$, we have $F_v=a_v$ and $\mathbf r_v=\mathbf h_v$, which establishes the base case. Suppose the expansion holds for every immediate predecessor $u\in\mathcal C(v)$. Substitution into Equation~\ref{eq:upstream} gives
\begin{align}
  \mathbf r_v
  &=\frac{a_v}{F_v}\mathbf h_v+
    \sum_{u\in\mathcal C(v)}\frac{F_u}{F_v}\mathbf D_{uv}\odot
      \left[\sum_{i\in S(u)}\frac{a_i}{F_u}
        \bigl(\mathbf K_{i\to u}\odot\mathbf h_i\bigr)\right]
        \nonumber\\
  &=\frac{a_v}{F_v}\mathbf h_v+
    \sum_{u\in\mathcal C(v)}\sum_{i\in S(u)}
      \frac{a_i}{F_v}
      \bigl(\mathbf D_{uv}\odot\mathbf K_{i\to u}\odot\mathbf h_i\bigr).
      \label{eq:appendix-induction}
\end{align}
The predecessor subtrees partition $S(v)\setminus\{v\}$, and the unique path satisfies
$\mathbf K_{i\to v}=\mathbf D_{uv}\odot\mathbf K_{i\to u}$ for $i\in S(u)$.
This proves Equation~\ref{eq:source-decomposition} by induction.

Equivalently, the scalar source coefficient telescopes along each path:
\begin{equation}
  \frac{a_i}{F_i}
  \prod_{(u,w)\in P(i,v)}\frac{F_u}{F_w}
  =\frac{a_i}{F_v}.
  \label{eq:appendix-telescoping}
\end{equation}
Path-product expansions also appear in tree state-space models \citep{xiao2024mambatree}. The area-normalized recurrence makes the geometric coefficient equal to the source's fraction of the target catchment, even after multiple confluences. The decomposition is an identity for the state-dependent gates realized in each forward pass.

\paragraph{Corollary 1: unit-retention reference.}
With $\mathbf D_e=\mathbf1$ on every edge, the decomposition reduces to
\begin{equation}
  \mathbf r_v^{(1)}
    =\frac{1}{F_v}\sum_{i\in S(v)}a_i\mathbf h_i.
  \label{eq:appendix-unit-retention}
\end{equation}
It preserves a constant source field: $\mathbf h_i=\mathbf h_\star$ for all $i$ implies $\mathbf r_v^{(1)}=\mathbf h_\star$. Unlike equal averaging at successive confluences, the source coefficient depends on contributing area rather than branching-dependent weight products.

\paragraph{Corollary 2: upstream-state amplitude.}
For learned retentions, each component of $\mathbf K_{i\to v}$ lies in $(0,1]$. If $\|\mathbf h_i\|_\infty\le B$ for all sources, then
\begin{equation}
  \|\mathbf r_v\|_\infty
  \le\sum_{i\in S(v)}\frac{a_i}{F_v}
     \|\mathbf K_{i\to v}\odot\mathbf h_i\|_\infty
  \le B\sum_{i\in S(v)}\frac{a_i}{F_v}=B.
  \label{eq:appendix-amplitude}
\end{equation}
The upstream-state bound is independent of the number of contributing sources.

\Needspace{20\baselineskip}
\subsection{Evaluation Order and Computational Cost}
\label{app:scan-computation}

Algorithm~\ref{alg:riverscan-evaluation} gives the spatial computation at one date, after local temporal encoding. The areas $F_v$ depend only on the tree and incremental areas and can be precomputed.

\begin{algorithm}[H]
\caption{RiverScan spatial evaluation at one date}
\label{alg:riverscan-evaluation}
\small
\begin{algorithmic}[1]
\Require Downstream tree $G=(V,E)$; local states $\mathbf h_v$, attributes $\mathbf s_v$, areas $a_v>0$; reach attributes $\mathbf e_{uv}$
\Ensure Accumulated states $\mathbf r_v$ and predictions $\widehat y_v$
\For{each $v$ in upstream-to-downstream topological order}
  \State $F_v\gets a_v+\sum_{u\in\mathcal C(v)}F_u$
  \State $\mathbf r_v\gets(a_v/F_v)\mathbf h_v$
  \For{each $u\in\mathcal C(v)$}
    \State Compute $\mathbf D_{uv}$ from completed $\mathbf r_u$ and $\mathbf e_{uv}$ using Equation~\ref{eq:retention}
    \State $\mathbf r_v\gets\mathbf r_v+(F_u/F_v)(\mathbf D_{uv}\odot\mathbf r_u)$
  \EndFor
\EndFor
\State Compute $\mathbf d_v$ in reverse topological order using Equation~\ref{eq:downstream}
\State Fuse $\mathbf h_v,\mathbf r_v,\mathbf d_v$ with attribute-conditioned gates, refine, and predict using Equations~\ref{eq:fusion}--\ref{eq:prediction}
\State \Return $\{\mathbf r_v,\widehat y_v\}_{v\in V}$
\end{algorithmic}
\end{algorithm}

Nodes and edges without mutual dependencies can be evaluated together; batch elements and dates are also independent within the spatial scan. With $H$ topological levels, the learned state dependence retains $H$ sequential stages. Standard selective sequence scans can precompute transitions from their sequence inputs \citep{gu2024}; RiverScan must form each source's accumulated state before evaluating its outgoing transition.

For fixed edge-feature dimension, dense state projections cost $O(|E|d^2)$ per evaluated date. Node projections and refinement of width $d_r$ cost $O(|V|d^2+|V|dd_r)$, giving
\begin{equation}
  O\!\left((|V|+|E|)d^2+|V|dd_r\right)
  \label{eq:appendix-core-cost}
\end{equation}
for the core spatial arithmetic. Elementwise weighting and aggregation alone cost $O(|E|d)$.

\section{Additional Experimental Details}
\label{app:experimental-details}

\subsection{Inputs, Chronological Splits, and Training}
\label{app:input-training}

For discharge in m$^3$\,s$^{-1}$ and area in km$^2$, define $c_{\mathrm{unit}}=86400\times10^3/10^6=86.4$, using $86400$ seconds per day, $10^6$ m$^2$ per km$^2$, and $10^3$ mm per metre. The runoff-depth conversions to mm\,day$^{-1}$ are
\begin{equation*}
  y_{v,t}=c_{\mathrm{unit}}\frac{Q_{v,t}}{F_v^{\mathrm{obs}}},\qquad
  q_v=c_{\mathrm{unit}}\frac{Q_v}{F_v},\qquad
  q_v^{\mathrm{loc}}=c_{\mathrm{unit}}\frac{Q_v^{\mathrm{loc}}}{a_v}.
\end{equation*}
Here $F_v^{\mathrm{obs}}$ is the reported drainage area (\texttt{area\_gov}), $F_v$ is the recursively accumulated area, and $a_v$ is the incremental area. The latter two conversions are used in Equation~\ref{eq:area-prior}.

\begingroup
\small
\setlength{\tabcolsep}{0pt}
\setlength{\LTcapwidth}{\linewidth}
\setlength{\LTpre}{6pt}
\setlength{\LTpost}{6pt}
\renewcommand{\arraystretch}{1.14}
\begin{longtable}{@{}>{\raggedright\arraybackslash}p{0.34\linewidth}@{\hspace{12pt}}>{\raggedright\arraybackslash}p{\dimexpr0.66\linewidth-12pt\relax}@{}}
\caption{Data and training settings (graph models unless stated otherwise).}
\label{tab:data-training}\\
\toprule
\textbf{Parameter} & \textbf{Value / setting} \\
\midrule
\endfirsthead
\caption[]{Data and training settings (continued).}\\
\toprule
\textbf{Parameter} & \textbf{Value / setting} \\
\midrule
\endhead
\midrule
\multicolumn{2}{r}{\footnotesize Continued on next page}\\
\endfoot
\bottomrule
\endlastfoot
\multicolumn{2}{@{}l}{\textbf{Dataset and inputs}}\\*
Dataset & LamaH-CE, Danube (LAMA-B) \\
River network & 608 gauges; 607 directed edges \\
Meteorological forcings & 10 daily variables per incremental catchment \\
Static attributes & Incremental area; mean elevation; mean slope \\
History length, $L$ & 365 days, ending on day $t$ \\
Prediction target & Same-day runoff depth (mm\,day$^{-1}$) \\
Observed discharge inputs & None \\
\addlinespace[4pt]
\multicolumn{2}{@{}l}{\textbf{Chronological splits}}\\*
Training & \mbox{1981-01-01} to \mbox{2000-09-30} \\*
Validation & \mbox{2000-10-01} to \mbox{2003-09-30} \\*
Test & \mbox{2003-10-01} to \mbox{2017-12-31} \\
\midrule
\multicolumn{2}{@{}l}{\textbf{Encoder and optimization}}\\*
Hidden-state dimension, $d$ & 256 \\
Output dropout & $0.4$ \\
Loss & NSE \\
Optimizer & Adam \citep{kingma2015} \\
Initial learning rate & $10^{-3}$ \\
Learning rate from epoch 15 & $5\times10^{-4}$ \\
Gradient clipping & Norm threshold of 1 \\
\midrule
\multicolumn{2}{@{}l}{\textbf{Batching and evaluation}}\\*
Batch size: graph models & 4 date windows, each with all 608 gauges \\
Batch size: full-catchment LSTM & 256 station-window pairs \\
Checkpoint selection & Lowest validation loss \\
Random seeds & 3 per configuration \\
\end{longtable}
\endgroup

\FloatBarrier
\subsection{Baseline Inputs and Connectivity Controls}

Table~\ref{tab:comparison-settings} specifies the model configurations and structural controls. Graph comparisons share incremental-catchment inputs and the LSTM architecture, with encoder weights learned separately in each run.

\begingroup
\small
\setlength{\tabcolsep}{0pt}
\setlength{\LTcapwidth}{\linewidth}
\setlength{\LTpre}{6pt}
\setlength{\LTpost}{6pt}
\renewcommand{\arraystretch}{1.14}
\newcommand{\ControlLabel}[1]{\multicolumn{2}{@{}>{\raggedright\arraybackslash}p{\dimexpr0.28\linewidth+10pt\relax}@{\hspace{10pt}}}{#1}}
\begin{longtable}{@{}>{\raggedright\arraybackslash}p{0.18\linewidth}@{\hspace{10pt}}>{\centering\arraybackslash}p{0.10\linewidth}@{\hspace{10pt}}>{\raggedright\arraybackslash}p{\dimexpr0.72\linewidth-20pt\relax}@{}}
\caption{Baseline configurations and controls. Depth denotes spatial layers. GraphSAGE uses no edge features in the depth study.}
\label{tab:comparison-settings}\\
\toprule
\textbf{Model} & \textbf{Depth} & \textbf{Configuration / reference} \\
\midrule
\endfirsthead
\caption[]{Baseline configurations and controls (continued).}\\
\toprule
\ControlLabel{\textbf{Control}} & \textbf{Implementation} \\
\midrule
\endhead
\midrule
\multicolumn{3}{r}{\footnotesize Continued on next page}\\
\endfoot
\bottomrule
\endlastfoot
\multicolumn{3}{@{}l}{\textbf{Spatial baselines}}\\*
GraphSAGE & 2 & Edge features included \citep{hamilton2017} \\*
GCN & 2 & \citet{kipf2017} \\*
GCNII & 2 & \citet{chen2020} \\*
GAT & 4 & \citet{velickovic2018} \\*
DAGNN & 2 & Stationwise prediction head \citep{thost2021} \\*
\addlinespace[3pt]
\multicolumn{3}{@{}l}{\textbf{Temporal references}}\\*
LSTM & --- & Full-catchment inputs; outside matched ablations \citep{kratzert2018} \\*
EA-LSTM & --- & Entity-aware LSTM \citep{kratzert2019} \\
\midrule
\ControlLabel{\textbf{Control}} & \textbf{Implementation} \\*
\midrule
\multicolumn{3}{@{}l}{\textbf{Topology controls}}\\*
\ControlLabel{Observed tree} & Retain the river-network connections \\*
\ControlLabel{Permuted tree} & Reassign stations; retain the unlabeled tree and station inputs/targets\newline
Recompute contributing areas and area fractions\newline
Use one permutation across all seeds \\*
\ControlLabel{No edges} & Remove all inter-station connections \\*
\ControlLabel{Fixed station groups} & Define headwater membership and GHI groups on the observed tree; hold fixed across comparisons \\
\end{longtable}
\endgroup

\begin{table}[!htbp]
\centering
\caption{Propagation depth on the observed Danube network (608 gauges). Station medians at validation-selected checkpoints are averaged over three seeds. GraphSAGE uses the plain variant. Bold marks the best depth per family and metric using unrounded values.}
\label{tab:depth}
\small
\setlength{\tabcolsep}{3pt}
\renewcommand{\arraystretch}{1.15}
\begin{tabular*}{\linewidth}{@{\extracolsep{\fill}}l*{8}{c}@{}}
\toprule
& \multicolumn{4}{c}{NSE med $\uparrow$} & \multicolumn{4}{c}{KGE med $\uparrow$} \\
\cmidrule(lr){2-5}\cmidrule(lr){6-9}
Family & 2L & 4L & 8L & 12L & 2L & 4L & 8L & 12L \\
\midrule
GraphSAGE & \textbf{0.705} & 0.679 & 0.652 & 0.647 & 0.749 & \textbf{0.757} & 0.707 & 0.735 \\
GCN & \textbf{0.684} & 0.669 & 0.645 & 0.642 & 0.755 & \textbf{0.762} & 0.736 & 0.721 \\
GCNII & \textbf{0.652} & 0.627 & 0.635 & 0.617 & \textbf{0.679} & 0.649 & 0.639 & 0.619 \\
GAT & 0.607 & \textbf{0.635} & 0.628 & 0.634 & 0.729 & \textbf{0.750} & 0.742 & 0.746 \\
\bottomrule
\end{tabular*}
\end{table}

\paragraph{Propagation depth.}
\label{app:depth-study}
Table~\ref{tab:depth} varies depth under the same checkpoint-selection and aggregation protocol. Three families decline from 2 to 12 layers; GAT peaks at 4 layers.

\FloatBarrier
\subsection{Metrics, Aggregation, and Bootstrap Intervals}
\label{app:evaluation-details}

NSE \citep{nash1970} and KGE \citep{gupta2009} are calculated from discharge after transforming it back to mm\,day$^{-1}$. For metric $m$ and seed $s$, the empirical CDF $\widehat F_{m,s}$ over a station group $V_g$ gives
\begin{equation}
  C_{m,s}^{(g)}=\int_0^1\widehat F_{m,s}^{(g)}(z)\,\mathrm dz
  =\frac{1}{|V_g|}\sum_{v\in V_g}
       \operatorname{clip}(1-m_{v,s},0,1).
  \label{eq:cdf-area}
\end{equation}
Lower CDF areas indicate better performance. All retained stations, including negative-score stations, remain included; each negative score contributes one, so its severity below zero is not distinguished. Table~\ref{tab:main} averages per-seed station medians and CDF areas over three seeds. The primary groups contain all 608 gauges, 314 headwaters with no upstream neighbour, and 294 non-headwaters.

Figures~\ref{fig:spatial-gain} and~\ref{fig:alignment} first average NSE across three seeds at each gauge.
\begin{equation}
  \overline{\mathrm{NSE}}_v^{(m)}
  =\frac{1}{3}\sum_{s=1}^{3}\mathrm{NSE}_{v,s}^{(m)},\qquad
  \Delta_v^{(m)}
  =\overline{\mathrm{NSE}}_v^{(\mathrm{obs})}
   -\overline{\mathrm{NSE}}_v^{(m)}.
  \label{eq:paired-gauge-nse}
\end{equation}
Figure~\ref{fig:alignment} reports the group median of seed-averaged NSE in (a), the group median of paired differences in (b), and the percentage of positive differences in (c). Averaging per-seed medians and taking a median after seed averaging need not agree. A median paired difference also need not equal the difference of two group medians. Each seed is evaluated at its validation-selected checkpoint.

GHI is defined recursively by
\begin{equation}
  \operatorname{GHI}(v)=
  \begin{cases}
    0,&\mathcal C(v)=\varnothing,\\
    1+\max_{u\in\mathcal C(v)}\operatorname{GHI}(u),&\text{otherwise}.
  \end{cases}
  \label{eq:gauge-hierarchy}
\end{equation}
The groups 0, 1, 2, 3, 4--6, and $\ge7$ have 314, 109, 61, 34, 46, and 44 gauges, respectively. GHI is the number of edges on the longest path from an upstream leaf to the target gauge.

In Figure~\ref{fig:alignment}, the gain over permutation is positive at roughly half of leaves, $96\%$ of gauges at GHI 4--6, and all 44 gauges at GHI $\ge7$. Median paired gains over the edge-free model rise from $+0.0277$ at leaves to $+0.1191$ at GHI $\ge7$. At observed-tree leaves, the upstream scan reduces to $\mathbf r_v=\mathbf h_v$; the full predictor also uses downstream context.

The pointwise $95\%$ intervals in Figure~\ref{fig:alignment}(b) use station-level bootstrap resampling within each group after seed averaging. This procedure does not account for dependence between connected gauges.

Table~\ref{tab:components} instead computes paired statistics within each of three matched seeds, then averages the seed summaries. For variant $m$ and station group $V_g$, the relative paired NSE change is
\begin{equation*}
  \Delta_{\mathrm{rel}}^{(m,g)}
  =\frac{100}{3}\sum_{s=1}^{3}
  \frac{
    \operatorname{median}_{v\in V_g}
    \bigl(\mathrm{NSE}_{v,s}^{(m)}-\mathrm{NSE}_{v,s}^{(\mathrm{full})}\bigr)
  }{
    \operatorname{median}_{v\in V_g}\mathrm{NSE}_{v,s}^{(\mathrm{full})}
  }.
\end{equation*}
Negative values favour RiverScan. This is a relative percentage, not a percentage-point difference. Win is the percentage of gauges where full RiverScan has higher NSE within each seed, averaged over the three seeds. Neither statistic is formed by first averaging each gauge's NSE across seeds, as in Figure~\ref{fig:alignment}. A paired median can be negative even when the variant's group median is higher, since taking a median and subtracting do not commute. Dashes mark the inapplicable full-model reference entries.

\paragraph{Upstream-state composition.}
\label{app:state-composition}
Figure~\ref{fig:state-composition} probes the completed upstream scan state $\mathbf r_v$ in Equation~\ref{eq:upstream}, before the projection $P_\uparrow$, static gate $\gamma_{\uparrow,v}$, and contextual fusion in Equation~\ref{eq:fusion}. The measurement therefore precedes refinement and the prediction head. This state summarizes encoded inputs over the full contributing area $F_v$; its coordinates are learned features with no prescribed storage or flux units. We suppress the date index below.

Split the state into $\mathbf r_v=\mathbf r_v^{\mathrm{up}}+\mathbf r_v^{\mathrm{local}}$, with $\mathbf r_v^{\mathrm{up}}=\sum_{u\in\mathcal C(v)}(F_u/F_v)\mathbf D_{uv}\odot\mathbf r_u$ and $\mathbf r_v^{\mathrm{local}}=(a_v/F_v)\mathbf h_v$. The first term combines completed tributary states after area weighting and selective retention; the second injects conditions over the local incremental region. For a nonzero state, the signed projection shares are
\begin{equation}
  p_{\mathrm{up},v}
    =\frac{\langle\mathbf r_v^{\mathrm{up}},\mathbf r_v\rangle}{\|\mathbf r_v\|_2^2},
  \qquad
  p_{\mathrm{local},v}
    =\frac{\langle\mathbf r_v^{\mathrm{local}},\mathbf r_v\rangle}{\|\mathbf r_v\|_2^2}.
  \label{eq:state-projection-shares}
\end{equation}
Each share is the scalar coefficient of a component's orthogonal projection onto the complete state: for example, $\operatorname{Proj}_{\mathbf r_v}(\mathbf r_v^{\mathrm{up}})=p_{\mathrm{up},v}\mathbf r_v$. Linearity gives $p_{\mathrm{up},v}+p_{\mathrm{local},v}=1$. Shares are dimensionless and can be negative or exceed one when components oppose one another. These coefficients describe the spatial composition of the completed upstream state before contextual fusion and prediction.

The physical reference is geometric: $F_{\mathrm{up}}=\sum_{u\in\mathcal C(v)}F_u$ (km$^2$) is the area represented by the incoming tributaries, and $t_v=F_{\mathrm{up}}/F_v=1-a_v/F_v$ is its fraction of the complete catchment. Panel (a) shows mean projection shares within upstream-area groups; panel (b) compares gauge-level shares with $t_v$. Projection shares combine geometric area weights with learned retention and the magnitudes and directions of regional encodings. Panel (b)'s logarithmic axes display only positive area fractions and shares, excluding leaf gauges whose upstream terms are zero.

\end{document}